\documentclass[preprint,12pt,3p]{elsarticle}

\usepackage{amssymb}
\usepackage{times}
\usepackage{epsfig}
\usepackage{graphicx}
\usepackage{amsmath}
\usepackage{subfigure}
\usepackage{algorithm}
\usepackage{algorithmic}
\usepackage{multirow}
\usepackage{color}
\usepackage{natbib}
\usepackage{caption}
\usepackage{mathptmx}  %the below two lines are used for mathcal font correction
\usepackage{mathrsfs}
\DeclareMathAlphabet{\mathcal}{OMS}{cmsy}{m}{n}
\DeclareSymbolFont{largesymbols}{OMX}{cmex}{m}{n}
\usepackage{amsthm}
\usepackage{amsfonts}
\usepackage{url} %2015-8-5
\usepackage{textcomp}
\usepackage[pagebackref=true,breaklinks=true,letterpaper=true,colorlinks,bookmarks=false,citecolor=blue]{hyperref}
\DeclareMathOperator*{\argmin}{argmin}
\DeclareMathOperator*{\argmax}{argmax}

\journal{arXiv}

\begin{document}

\begin{frontmatter}

\title{Hierarchical Spatial-aware Siamese Network for Thermal Infrared Object Tracking}
\author[label1]{Xin Li\corref{equ}}
\author[label1]{Qiao Liu \corref{equ}}
\address[label1]{School of Computer Science and Technology, Harbin Institute of Technology Shenzhen Graduate School, China}
\author[label1]{Nana Fan}
\author[label1]{Zhenyu He\corref{cor1}}\ead{zhenyuhe@hit.edu.cn}
\author[label2]{Hongzhi Wang\corref{cor1}}\ead{wangzh@hit.edu.cn}
\address[label2]{School of Computer Science and Technology, Harbin Institute of Technology, Harbin, China}
\cortext[equ]{Contribution equally}

\cortext[cor1]{I am corresponding author}

\begin{abstract}
Most thermal infrared (TIR) tracking methods are discriminative, {treating} the tracking problem as a classification task. However, the objective of the classifier (label prediction) is not coupled to the objective of the tracker (location estimation). The classification task focuses on the between-class difference of the arbitrary objects, while the tracking task mainly deals with the within-class difference of the same objects. In this paper, we cast the TIR tracking problem as a similarity verification task, which is coupled well to the objective of the tracking task. We propose a TIR tracker via a Hierarchical Spatial-aware Siamese Convolutional Neural Network (CNN), named \textbf{HSSNet}. To obtain both spatial and semantic features of the TIR object, we design a Siamese CNN that coalesces the multiple hierarchical convolutional layers. Then, we propose a spatial-aware network to enhance the discriminative ability of the coalesced hierarchical feature. Subsequently, we train this network end to end on a large visible video detection dataset to learn the similarity between paired objects before we transfer the network into the TIR domain. Next, this pre-trained Siamese network is used to evaluate the similarity between the target template and target candidates. Finally, we locate the {candidate that is most similar to the tracked target}. Extensive experimental results on the benchmarks VOT-TIR 2015 and VOT-TIR 2016 show that our proposed method achieves favourable performance {compared to} the state-of-the-art methods.
\end{abstract}


\end{frontmatter}

%%
%% Start line numbering here if you want
%%
% \linenumbers

%% main text
\section{Introduction}
\label{introduction}
In recent years, both the price and the size of thermal cameras have decreased, while the resolution and quality of thermal images have improved, which has opened up new application areas such as surveillance, rescue, and driver assistance at night~\cite{TC,TIR2016review}. Thermal infrared (TIR) object tracking is often used as a subroutine that plays an important role in these vision tasks. It has several superiorities over visual object tracking~\cite{zhang2017robust,zhang2015robust,ma2018visual,zhang2015single,zhang2015hybrid,sui2015robust,liu2017visual,PatchTracking,sparsetracking,gao2018high,MOT,wang2018object}. For example, TIR tracking is not sensitive to variation of the illumination, whereas visual tracking usually fails in poor visibility. In addition, it can protect privacy in some real-world scenarios such as surveillance of private places.

Despite its many advantages, TIR object tracking faces a lot of challenges. First, TIR objects have several adverse properties, such as the absence of visual colour patterns, low resolution, and blurry contours. {Also, TIR objects often lie on} a complicated background that has dead pixels, blooming, and distractors~\cite{ABCD}. These adverse properties hinder the extraction of discriminative features by the feature extractor, which severely degrades the quality of the tracking model. Second, several other challenges are faced by TIR tracking, such as deformation, occlusion, and scale variation. In order to handle these challenges, several TIR trackers have been proposed over the past years. For instance, Li \emph{et al.}~\cite{li2014real} propose a TIR tracker based on sparse theory and compressive Harr-like features which can handle the occlusion problem to some extent. Gundogdu \emph{et al.}~\cite{TBOOST} use multiple correlation filters (CFs) with the histogram of oriented gradient features to construct an ensemble TIR tracker which can adapt the changed appearance of the object due to the proposed switching mechanism. To alleviate the fact that a single feature is not robust to the various challenges, in~\cite{gao2016infrared}, the authors present a sparse representation-based TIR tracking method using the fusion of multiple features. However, these methods do not solve the challenges of TIR tracking well because it is difficult to obtain the discriminative information of the TIR objects using these hand-crafted features.

Considering the powerful representation ability of the Convolutional Neural Network (CNN), some works~\cite{CNN-CF,liu2017deep}  introduce CNN features into TIR tracking. Unfortunately, these trackers have not made great progress for several reasons. First, the used CNN feature is obtained from a classification network, which is not optimal for tracking because the objectives of these two tasks are not explicitly coupled with the network learning. Second, these trackers are not robust to various challenges since they only use the feature of a single CNN layer. Third, the network is only trained on limited TIR images, {so this training} is insufficient to obtain a robust feature.

%The classification task focuses on the difference of the between-class of the arbitrary objects, while the tracking task mainly cares about the difference of the within-class of the same ones.
%For example, given a video frame which has several persons and cars. To the classification task, it focuses on how to discriminate the persons and cars, while the tracking task mainly cares about how to distinguish the different cars and locate the object car in the next frame.

Most recently, by casting the tracking problem as a similarity verification problem, a visual tracker, Siamese-fc~\cite{Siamese-fc}, has been proposed. It simply uses a pre-trained Siamese network as a similarity function to verify whether the target candidate is the tracked target in the tracking process. Compared to the classification network, the pre-trained Siamese network is more coupled to the tracking task. Thus, the feature extracted from the pre-trained Siamese network has more discriminative ability. In this paper, we use a Siamese network to carry out TIR tracking. However, there are several problems that must be solved. First, this Siamese network uses the feature of the last convolutional layer, which is not robust for TIR tracking due to the adverse properties of the TIR objects. Second, {a dataset with sufficient public TIR images to train the network is lacking}.

To address these problems, we propose a TIR tracker using a hierarchical spatial-aware Siamese CNN. Specifically, to obtain richer spatial and semantic features for TIR tracking, we first design a Siamese CNN that coalesces the deep and shallow hierarchical convolutional layers, since the TIR tracking needs not only the deep-level semantic features to distinguish the different objects but also the shallow-level spatial information to precisely locate the target object. Subsequently, to improve the discriminative ability of the coalesced hierarchical features, a spatial-aware network is integrated into the Siamese CNN. Additionally, we suggest that the deep features learned from visible images can also represent the TIR objects. Therefore, to handle the lack of training data, we train the proposed Siamese network on a large visible video dataset to learn the similarity between paired objects before we transfer the learned network into the TIR domain. Then, the learned Siamese network is used to evaluate the similarity between the target template and target candidates. Finally, we locate the candidate most similar to the tracked target without any adaptation in the tracking process. The experimental results show that our method achieves satisfactory performance.

%contributions
The contributions of the paper are three-fold:
\begin{itemize}
  \item We propose a simple yet effective hierarchical convolutional features fusion method, which can obtain richer spatial and semantic feature representation of the TIR object.
  \item A spatial-aware network is designed to enhance the discriminative ability of the coalesced hierarchical features.
  \item We carry out extensive experiments on benchmark datasets to show that our proposed TIR tracker performs favorably against state-of-the-art methods.
\end{itemize}

The rest of the paper is organized as follows. Section~\ref{Related Work} introduces the most closely related works briefly. Section~\ref{PA} describes the main part of the proposed approach. Section~\ref{experiment} carries out the experiment and presents the results while Section~\ref{Conclusion} draws a short conclusion.

\section{Related Work}
\label{Related Work}
In this section, we discuss two classes of the most closely related works: classification-based trackers and verification-based trackers. We first review the classification-based TIR trackers and analyse the {drawbacks. Then, we discuss the verification-based trackers and state the advantages.}

\vspace{3mm}

\noindent{\textbf {Classification-based trackers.}} These trackers have received more attention in TIR object tracking. To deal with various challenges, a variety of the classification-based TIR trackers are presented based on sparse representation~\cite{li2014real,gao2016infrared,Wen2018Inter,wen2018robust}, multiple instances learning~\cite{shi2013infrared}, kernel density estimation~\cite{liu2012infrared}, low-rank sparse learning~\cite{he2016infrared,Jie2018Adaptive}, structural support vector machine~\cite{Yu2017Dense}, correlation filter~\cite{TBOOST,he2015infrared,asha2017robust}, and deep learning~\cite{CNN-CF,liu2017deep,li2018fusing}. For instance, to address the occlusion problem, He et al.~\cite{he2016infrared} propose a robust low-rank sparse tracker using the low-rank constraints to capture the underlying structure of the TIR object. {In~\cite{Yu2017Dense}, the authors suggest that the structure learning is more suitable for the object tracking task. Therefore, they design an online dense structural learning TIR tracker via a structural support vector machine. To overcome the training time-consumptiion problem of training, they optimize the LaRank algorithm by using Fast Fourier Transform.}  In~\cite{TBOOST}, the authors propose an ensemble correlation filter TIR tracker which can handle the variation in appearance due to the proposed switch mechanism. In order to extract more discriminative features of the TIR object, Liu et al.~\cite{liu2017deep} investigate the deep convolutional feature for TIR object tracking. They propose a Kullback-Leibler divergence based fusion tracker by exploiting multiple convolutional layer features. {In addition, the classification-based methods are often used in RGB and thermal (RGB-T) tracking and visual tracking. For example, Li et al.~\cite{li2018fusing} propose a two-stream fusion network to solve the RGB-T tracking. They use two-stream fusion structure to obtain the complementary information from RGB and thermal sources. Yun et al.~\cite{yun2018multi} use a multi-layer CNN and an important region selection strategy to solve the visual tracking problem.}
Although these classification-based trackers achieve the promising performance, the objective of these classifiers is still not coupled to the objective of the tracking task.
The goal of the classifier is to predict the class label of a sample that usually is used in pattern recognition~\cite{you2014local,zhu2015adaptive,jing2017super,ge2017structure,chen2006two,ou2016multi,ou2014robust,guo2016robust,shi2016two,lai2014multilinear,yan2018insulator,lai2016approximate,yi2017unified,JointPCA}, while the aim of the tracker is to estimate object position accurately. None of these classification-based TIR trackers can take into account the within-class difference, while the tracking task mainly considers it. {Therefore, we suggest that the classification-based trackers are not optimal.} Unlike these classification-based trackers, in this paper, we cast the tracking task as a similarity verification problem, which is more coupled to the tracking task.
\vspace{3mm}

\noindent{\textbf{Verification-based trackers.}} {The similarity verification task focuses on comparing two arbitrary objects to determine whether they are the same or not.} Unlike the classification task, it does not care about between-class differences but considers within-class differences. Therefore, we suggest that it is more suitable for the tracking task.
Most recently, verification-based trackers have been presented in  object tracking with competitive results. This kind of method is often based on a Siamese architecture, which consists of two identical or asymmetric sub-networks joined at their outputs.
For instance, YCNN~\cite{YCNN} learns discriminating features of growing complexity while simultaneously learning the similarity between the template and search region with corresponding prediction maps using a shallow Siamese network.
Tao \emph{et al.}~\cite{SINT} propose a Siamese invariance network to learn a generic matching function for tracking (SINT). The learned matching function is used to match the initial target with candidates and returns the candidate most similar to the tracked target.
Bertinetto \emph{et al.}~\cite{Siamese-fc} present a fully-convolutional Siamese network (Siamese-fc) to learn the similarity function of two arbitrary objects for tracking. However, it often fails when the appearance of the target changes drastically due to the lack of the updating strategy.
Subsequently, Valmadre \emph{et al.}~\cite{CF-NET} combine the correlation filter with Siamese architecture to construct a deep correlation filter learner (CFNet), which explains the correlation filter as a differentiable layer in a deep neural network.
However, most of these Siamese networks are not compatible with TIR tracking, because most of them use the single convolutional layer feature to represent the object which is not robust to the TIR tracking.
To adapt the TIR tracking, in our Siamese network, we design a hierarchical Siamese network which coalesces multiple hierarchical convolutional layers to obtain richer spatial and semantic features. Furthermore, we also design a spatial-aware network to improve the discriminative ability of the coalesced hierarchical features.

\section{Hierarchical Spatial-aware Siamese Network}
\label{PA}
In this part, we first give the overall framework in Section~\ref{AO} and then present the hierarchical spatial-aware Siamese network architecture in Section~\ref{SNA}. Next, we explain how to train this network to learn the similarity function in Section~\ref{TSN},
and finally we present the tracking interface in Section~\ref{TI}.

\subsection{Algorithmic Overview}
\label{AO}

\begin{figure}[ht]
\begin{center}
\includegraphics[width=1\textwidth]{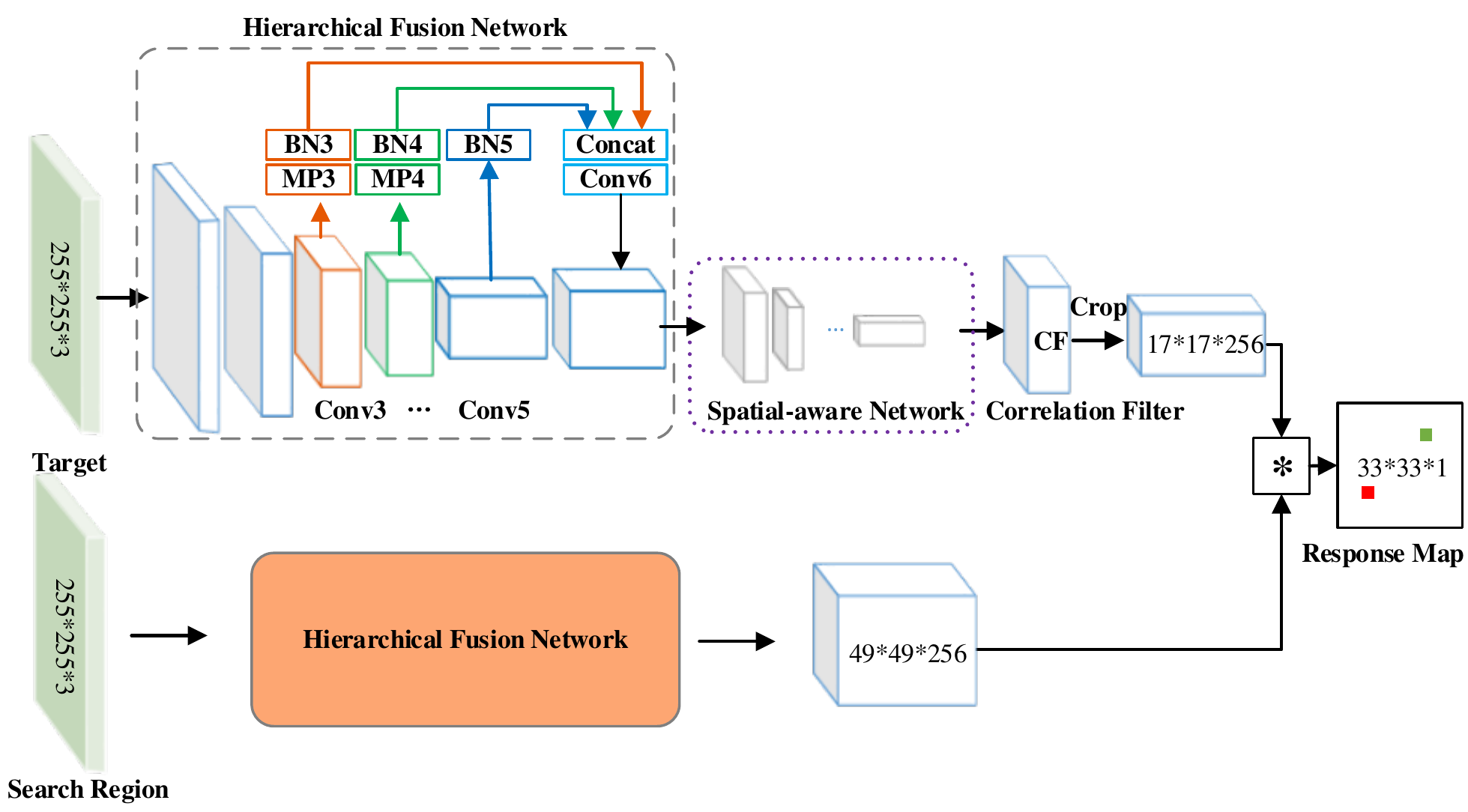}
\end{center}
\caption{The architecture of the proposed Siamese network (HSSNet). The hierarchical spatial-aware network coalesces multiple convolution layers and a spatial-aware network. MP, BN, Concat, Conv, Crop, and CF denote the max pooling layer, batch normalization layer, concat layer, convolution layer, crop layer, and correlation filter layer, respectively. The \textbf{\textcolor{green}{green} and \textcolor{red}{red}} pixels in the response map represent the similarity score between the target template and candidate. The pixel with the highest score is regarded as the final tracked target.}
\label{architecture}
\end{figure}

The proposed method is based on a hierarchical Siamese network that coalesces multiple hierarchical convolutional layers and a spatial-aware network, as shown in Fig.~\ref{architecture}. This network consists of two asymmetric branches, which consist of two shared hierarchical fusion networks and a spatial-aware network, finally joined by a cross-correlation layer. The output of the Siamese network is a response map which denotes the similarity between the multiple candidates and the target template.

In the tracking stage, our goal is to use the pre-trained Siamese network to locate the tracked target.
It can be simply formulated as the following cross-correlated operator:
\begin{equation}\label{similarityfunction}
  f(x,z)=\omega(\nu(\varphi(x)))\ast \varphi(z),
\end{equation}
where $\varphi(\cdot)$ is an embedding function and denotes the learned Siamese hierarchical fusion network in Fig.~\ref{architecture}. The spatial-aware block $\nu(\varphi(x))$ scales the fusion feature map adaptively. The CF block $ w=\omega(\nu(\varphi(x)))$ computes a standard CF target template $w$ from the feature map $\nu(\varphi(x))$ by solving a ridge regression problem in the Fourier domain.
As shown in Fig.~\ref{architecture}, we just need to input a target and a search region, the Siamese network will return a response map that measures the similarity between the candidates and the target template. {After that}, we choose the corresponding candidate with the maximal value of the response as the final tracked target, and then the coordinates are mapped into the original frame to locate the position of the target.

\subsection{The Network Architecture}
\label{SNA}
The proposed Siamese network is composed of two asymmetric branches, as shown in Fig.~\ref{architecture}. For individual branches, inspired by AlexNet~\cite{AlexNet}, we design a deep network architecture, which consists of several types of layers commonly used in CNN.
%The details of these layers are shown in Table~\ref{structuredetails}.
In the following, we mainly introduce the distinctive designs of the proposed networks.

\vspace{3mm}
%max pooling
\noindent{\textbf{Max pooling.}} Location information of the object is not important for the classification task but is required for the tracking task. After the max pooling layer, the feature map often loses the location information of the object to some extent.  Like AlexNet~\cite{AlexNet}, which has two max pooling layers, our network also has two max pooling layers at the early stage to reserve more location information. On the other hand, the max pooling is robust to the local noises  because it introduces the invariance to the local deformation. Therefore, it is important for the tracking task since the tracked object changes its appearance over time.
\vspace{3mm}

\noindent{\textbf{Batch normalization.}} To accelerate the training of the Siamese network, we add a batch normalization layer~\cite{BN} after each convolutional layer. The effectiveness of batch normalization has been shown in many deep networks. Additionally, in the fusion part of the network, we normalize multiple hierarchical convolutional feature maps using the batch normalization operator and then combine them into one single output cube.
\vspace{3mm}

%multi-layers reasons
\noindent{\textbf{Hierarchical fusion network.}} {This differs from the previous Siamese architecture}, which just uses the feature from the last layer to represent the object. However, the last layer feature lacks the spatial information, which is not robust for TIR tracking. To obtain more robust features for TIR tracking, our proposed Siamese network coalesces multiple hierarchical convolutional layers, since we note that the tracking task not only needs the discriminative semantic information of the deep layers to distinguish the different objects but also needs the spatial information of the shallow layers to precisely locate the target position.
%how to do
In order to coalesce these hierarchical convolutional layers, which have a different spatial resolution, we exploit the max pooling to downsample the shallow convolutional layer to the same resolution as the deep convolutional layer.  Before concatenating these translated feature maps, we adopt the batch normalization layer to normalize these feature maps because we want to balance the influence of these feature maps.
Three hierarchical convolutional feature maps $f^{3}\in\mathbb{R}^{53\times53 \times 384}$, $f^{4}\in\mathbb{R}^{51\times51 \times 384}$, and $f^{5}\in\mathbb{R}^{49\times49 \times 256}$, which denote the features of Conv3, Conv4, and Conv5, respectively, are considered.
Each of these three feature maps has a different spatial resolutions. To fuse these feature maps, we define two functions, $mp(\cdot)$ and $bn(\cdot)$, to represent the max pooling and batch normalization operator, respectively. Thus, the fused feature map can be formulated as follows:
\begin{equation}\label{fusedmap}
  fmap=concat(bn(mp(f^{3})),bn(mp(f^{4})),bn(mp(f^{5}))),
\end{equation}
where $concat(\cdot,\cdot)$ denotes the concat layer,  which concatenates the multiple feature maps in the channel direction. After that, we find that the dimension of the fused feature map $fmap\in\mathbb{R}^{49\times49 \times 1024}$ is too high to train the network quickly. Therefore, it is necessary to reduce the dimension of the fused feature map. In our network, we use a $1\times 1$ convolutional layer to reduce the channel dimension of the fused feature map, as shown for Conv6 in Fig.~\ref{architecture}. This convolutional layer not only reduces the dimension of the feature map but also assigns the weights to different hierarchical convolutional layers adaptively. The final fused hierarchical convolutional feature map, $finalmap$, can be formulated as follows:
\begin{equation}\label{finalfusedmap}
  finalmap=conv(fmap),
\end{equation}
where $conv(\cdot)$ denotes a $1\times1$ convolutional operator and  $finalmap\in\mathbb{R}^{49\times49 \times 256}$ has a suitable dimension for training.
\vspace{3mm}

\begin{figure}[htbp]
\begin{center}
\includegraphics[width=1\textwidth]{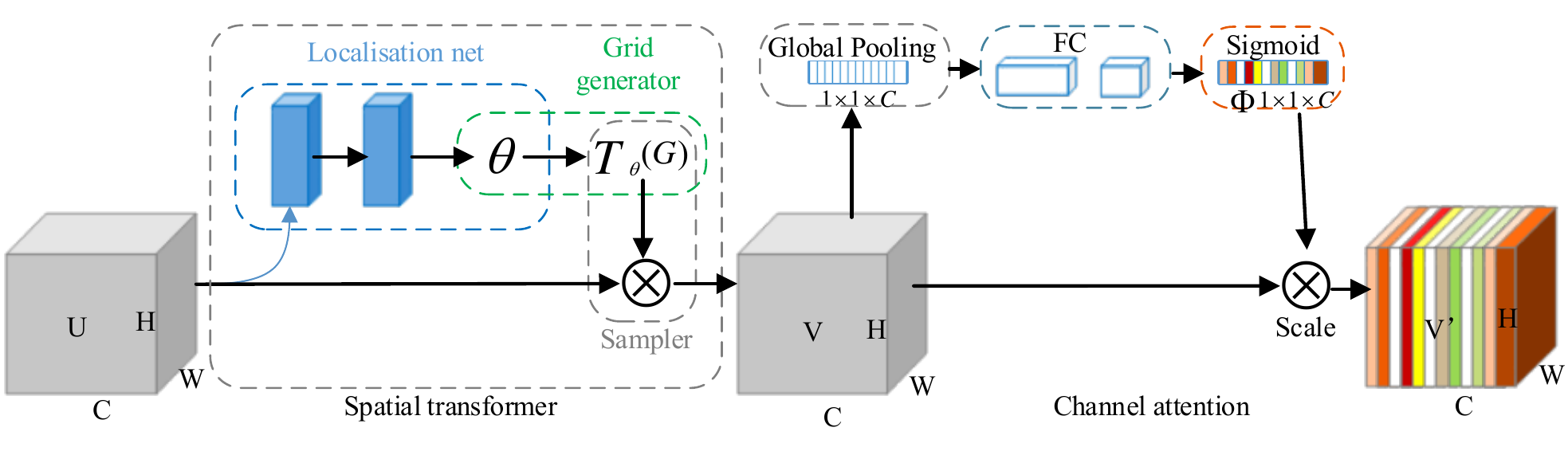}
\end{center}
\caption{The architecture of the spatial-aware network, which consists of a spatial transformer network and a channel attention network.}
\label{STAN}
\end{figure}
\noindent{\textbf{Spatial-aware network.}}
{When} we get the fused hierarchical feature using Eq.~\ref{finalfusedmap}, it contains spatial and semantic information simultaneously. However, the fused feature is still not robust to the spatial rotation, scaling, and translation, which are important for the tracking task. Furthermore, the different feature channels make the same contribution in our task, which is not reasonable. To address these problems, we design a spatial-aware network which is cascaded by two sub-networks: a spatial transformer network and a channel attention network, as shown in Fig.~\ref{STAN}.

\vspace{3mm}
\noindent{\textbf{Spatial transformer.}} The goal of the spatial transformer network is to learn a 2D affine transformation which includes rotation, scaling, and translation. Then, the learned spatial transformer is applied to the input feature map. Spatial Transform Network (STN)~\cite{STN} has been proven to be effective for the many vision tasks and is also suited well to our application. Therefore, we decided to exploit STN in our network to achieve {spatial transformability}. STN has three components,  namely the localization net, {grid generation}, and sampler, as shown in the spatial transformer in Fig.~\ref{STAN}. The localization net receives the input feature map $U$ and returns the corresponding transformation parameters $\theta$ via several hidden layers. We use three convolutional layers and one fully connected layer as the hidden layers.  Here, $\theta$ is a 2D affine transformation with six-dimensions. The grid generator maps the transformation into the input feature map. The process can be formulated as follows:
\begin{equation}\label{maping}
  \begin{pmatrix} x^{in} \\ y^{in} \end{pmatrix}=T_{\theta}(G)=\begin{bmatrix} \theta_{11} & \theta_{12} & \theta_{13} \\ \theta_{21} & \theta_{22} & \theta_{23} \end{bmatrix}\begin{pmatrix} x^{out} \\ y^{out} \\ 1 \end{pmatrix},
\end{equation}
where $(x^{out},y^{out})$ are the coordinates of the regular grid in the output feature map $V$, and $(x^{in},y^{in})$ are the coordinates in the input feature map $U$.
The sampler takes the set of sampling points $T_{\theta}(G)$ with the input feature map $U$, and then produces the sampled output feature map $V$. More details are presented in \cite{STN}.

\vspace{3mm}
\noindent{\textbf{Channel attention.}} Since the feature map $V \in  \mathbb{R}^{W\times H\times C}$ has multiple channels, each channel has a certain type of visual pattern. Therefore, it is not reasonable to treat all feature channels as having the same weight. To identify the more important feature channel for our task, we use a channel attention network to re-weight each channel of the feature map. Here, we adopt an SE-block~\cite{SENet} as our channel attention network. The SE-block is used in the classification network and is proven to be effective. It includes a global pooling layer, two fully-connected layers, and a Sigmoid activation layer. Given the Sigmoid layer output of the attention network $\Phi \in \mathbb{R}^{1\times1\times C}$, the final re-weighted feature map $V^{'}\in  \mathbb{R}^{W\times H\times C}$ is calculated by:
\begin{equation}\label{SEscale}
  V^{'}=scale(V,\Phi),
\end{equation}
where $scale(\cdot,\cdot)$ is a channel-wise multiplication function.

\vspace{3mm}

%CF layer
\noindent{\textbf{Correlation filter.}}
As in CFNet~\cite{CF-NET}, CF is interpreted as a differentiable CNN layer in our Siamese architecture. So, the errors can be propagated through the CF layer back to the CNN features and the overall Siamese network can be trained end to end. In addition, the CF layer can be used to update the target template in the tracking process. The dynamic target template can adapt to the variation in the appearance of the target more flexibly.
%In this paper, we use the CF layer to learn a dynamic target template to adapt the appearance variation of the target.

\subsection{Training the Network}
\label{TSN}
{In order to train a general similarity verification function that can evaluate the similarity of a pair of objects, we need a large-scale annotated video dataset. Given the limited scale of the existing TIR tracking and detection dataset, we choose the RGB detection video dataset: ILSVRC2015 because we believe that TIR objects and RGB objects have something in common. ILSVRC2015 has more than 4000 videos containing more than 2 million labelled bounding boxes. The scale of the dataset is sufficient for our task. Once the proposed network has been learned from the ILSVRC2015 dataset, we transfer it into the TIR domain.}
\vspace{3mm}

\noindent{\textbf{{Generation of Training Samples.}}} As shown in Fig.~\ref{architecture}, the proposed Siamese network needs a target and a search region as the inputs. The output measures the similarity between a target and multiple candidates. In order to train the network, we prepare the training pairs (a target and a corresponding search region) from a large visible images video detection dataset from ImageNet~\cite{ImageNet} and the corresponding labels as in~\cite{Siamese-fc}. First, we scale the original image frame with a scale factor $s$. Then, we crop a target with a fixed size  and a search region that is centered  on the target at every frame of the video. Finally, we randomly choose a target and a search region within an interval of $T$ frames  in the same video as a training pair. Assuming that the bounding box size of the target is $(w,h)$, and the cropped size is $A$, the scale factor $s$ can be formulated as:
\begin{equation}\label{crop}
  s(w+2p)\times s(h+2p)=A,
\end{equation}
where $p=\frac{(w+h)}{4}$ is the padding context margin. Given the response map $\mathcal{D} \in \mathbb{R}^{2}$ of the network, we suggest that an element $u\in \mathcal{D}$ is a positive sample if it is within radius $R$ of the center
\begin{equation}\label{crorespondinglabel}
y[u]= \begin{cases}
+1 & \text{if } k \|u-c \| \leq R \\
-1 & \text{otherwise, }
\end{cases}
\end{equation}
where $k$ denotes the total stride of the network.  For all training pairs, the corresponding labels are calculated by Eq.~\ref{crorespondinglabel}.
\vspace{3mm}

\noindent{\textbf{Loss function.}} We add a logistic loss layer to train the network at the end of the Siamese network
\begin{equation}\label{logisticloss}
  \ell(y,v)=\log(1+exp(-yv)),
\end{equation}
where $v$ denotes the real score of a single target-candidate pair returned by the model.  $y \in \{+1, -1\}$ represents the ground-truth label of this pair. For the loss of the response map, which measures the similarity between a target and multiple candidates, the mean of the individual losses is exploited:
\begin{equation}\label{meanloss}
  L(y,v)=\frac{1}{\mid \mathcal{D} \mid}\sum_{u \in \mathcal{D}}\ell(y[u],v[u]).
\end{equation}
The parameters $\theta$ of the Siamese network can be obtained by minimizing the loss function
\begin{equation}\label{minloss}
\argmin_{\theta} \mathbb{E}_{z,x,y} L(y,f(z,x;\theta)).
\end{equation}
To solve problem~\ref{minloss}, we can use Stochastic Gradient Descent (SGD) to solve it.

\subsection{Tracking Interface}
\label{TI}
Once the similarity function $f$ has been learned, we simply exploit it as a prior in the tracking without any adaptation. We use a simple strategy to verify the target candidates. Given the cropped target $x_{t-1}$ at the $(t-1)$-th frame and a search region $z_{t}$ at the $t$-th frame, the tracked target at the $t$-th frame can be calculated by the following formulation:
\begin{equation}\label{tracking}
  \hat{z}_{t,i}=\argmax_{z_{t,i}} f(z_{t},x_{t-1}),
\end{equation}
where $z_{t,i} \in z_{t}$ is the $i$-th candidate in the search region $z_{t}$.

\vspace{3mm}
\noindent{\textbf{Scale estimation.}} In order to enhance the accuracy of the tracking, we adopt a simple but effective scale estimation method~\cite{Multi-viewCF}.  The search regions of three different scales are inputted in the network for comparison with the target, and the maximum response map and the corresponding scale are returned.

\section{Experiments}
\label{experiment}
To demonstrate the effectiveness of our approach, we conduct the experiments on two TIR tracking benchmarks. First, we give the implementation details in Section~\ref{details} and describe the evaluation criteria in Section~\ref{criterion}. Then, we carry out an ablation experiment in Section \ref{experiment1} and an external comparison experiment in Section~\ref{experiment2}, respectively.

\subsection{Implementation Details}
\label{details}
\noindent{\textbf{Training.}} We train the proposed Siamese network on the large video detection dataset (ILSVRC2015) from ImageNet~\cite{ImageNet} by solving Eq.~\ref{minloss} with straightforward SGD using MatConvNet~\cite{matconvnet}. {In the training samples generation stage (see the part on training samples generation in Section~\ref{TSN}), we set the interval $T$ to $100$ and the radius $R$ to 8 pixels.}  The training is performed over $40$ epochs, {each consisting of by $50,000$ sampled pairs. The initial parameters of the network follow a Gaussian distribution by using the improved Xavier method~\cite{matconvnet}.} We use mini-batches with a size of $8$ to calculate gradients in each iteration. The learning rate is annealed geometrically in each epoch from $10^{-2}$ to $10^{-5}$. { Most of these hyper-parameters are the same as those in~\cite{Siamese-fc}.} After finishing the training, the result of the $37$-th epoch is exploited as the final model.
\vspace{3mm}

\noindent{\textbf{Tracking.}} The experiments are conducted in MATLAB 2015b  with a GTX 1080 GPU card.  In Algorithm~\ref{alg:HSNet}, we give the main steps of the proposed tracking algorithm.  In order to deal with the scale variation, we use three fixed scales $\{0.9745, 1, 1.0375\}$ to search the object. The scale is updated by linear interpolation with a factor of $0.59$ to provide damping. The average frame rate of the proposed tracker is $10$ frames per second (fps).

\begin{algorithm}[h]
 \caption{The proposed tracker (HSSNet) }
  \label{alg:HSNet}
  \begin{algorithmic}[1]
  \STATE \textbf{Inputs}: initial target state $x_{1}$, the learned similarity function $f$ using Eq.~\ref{minloss}.
   \STATE \textbf{Outputs}: the estimated target state $\hat{x}_{t}$.
   \WHILE{$t< length(sequence)$}
    \STATE Crop three different scales' search regions $z_{t}^{1}, z_{t}^{2},z_{t}^{3}$ based on the target state $x_{t-1}$.
    \STATE Calculate the optimal target state using Eq.~\ref{tracking} and return the best scale factor.
   \ENDWHILE
  \end{algorithmic}
\end{algorithm}

\subsection{Evaluation Criteria}
\label{criterion}
Accuracy (A) and robustness (R) are adopted as the performance measures due to their high interpretability~\cite{VOTperformance1}.
A can be calculated using the following formulation:
\begin{equation}\label{accuracy}
  A =\frac{1}{n}\sum_{t=1}^{n}\frac{\parallel B_{t} \cap G_{t}\parallel}{\parallel B_{t}\cup G_{t}\parallel},
\end{equation}
where $B_{t}$ denotes the area of the predicted bounding box, $G_{t}$ denotes the area of ground-truth at the $t$-th frame, and $n$ is the frame number of the dataset.
The robustness counts the number of failures. The tracker is failing when $B_{t} \cap G_{t}$ is lower than a given threshold.
The tracking results are often visualized by an A-R raw plot and an A-R ranking plot~\cite{ARplot}.

In addition, to predict the overall performance of the tracker, A and R are integrated into the expected average overlap (EAO) measurement. The EAO curve and EAO score are often used to evaluate the overall performance~\cite{VOT2015}.

%Apart from accuracy and robustness, the tracking speed is also an important property that indicates the practical usefulness of trackers in the particular applications. To reduce the influence of hardware, the VOT-TIR 2016~\cite{VOT-TIR2016} adopts a new unit for reporting the tracking speed as in~\cite{VOT2014}, called equivalent filter operations (EFO) that report the tracker speed in terms of a predefined filtering operation that the toolkit automatically carries out prior to running the experiments.

\subsection{Ablation Experiments on VOT-TIR 2015}
\label{experiment1}
In this section, we show the effectiveness of each component of our tracker. An internal comparison experiment  is conducted on the benchmark VOT-TIR 2015~\cite{VOT-TIR2015}.
%We use the accuracy and robustness as the evaluation criteria to evaluate the performance of the trackers and exploit the A-R raw plot and A-R ranking plot to visualize the experiment results.
\vspace{3mm}

\noindent{\textbf{Datasets.}} VOT-TIR 2015 has twenty TIR sequences and each sequence has several local attributes, such as dynamics change, occlusion, camera motion, and object motion. The tracker's performance on these attributes is often compared with others.
\vspace{3mm}

\noindent{\textbf{Compared trackers.}} First, to demonstrate that our hierarchical convolutional layer fusion method is effective, we compare the tracker HSNet with HSNet-lastlayer, where the HSNet tracker uses the hierarchical convolutional layer fusion method, while HSNet-lastlayer just uses the last single convolutional layer.
Then, HSNet, HSNet-ST (HSNet with spatial transformer), and HSNet-CA (HSNet with channel attention) are compared to check the effectiveness of each part. Furthermore, we compare the HSSNet with HSNet to show that the overall spatial-aware network can enhance the tracking performance.
Finally, to demonstrate that the scale estimation strategy is also effective, we compare HSSNet with HSSNet-noscale, which is HSSNet without the scale estimation strategy. %Futhermore, we also compare Siame

\vspace{3mm}
\begin{figure}[ht]
\centering
\includegraphics[width=1\textwidth]{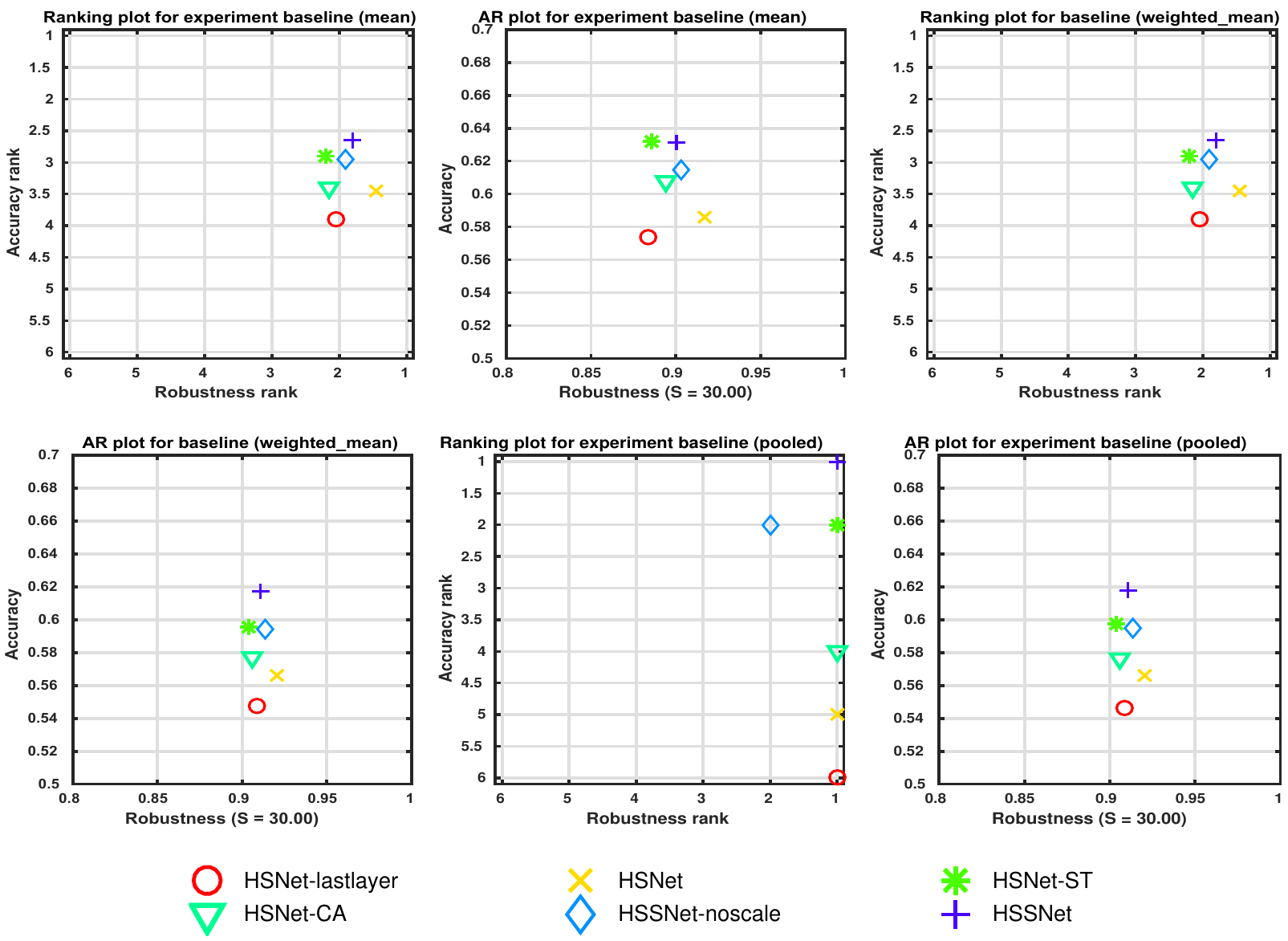}
\caption{A-R ranking plots and A-R raw plots on VOT-TIR 2015. The better the performance a tracker obtains, the closer to the top right of the plot it is plotted.}
\label{fig:VOT-TIR2015AR-ranking}
\end{figure}

\noindent{\textbf{Evaluation and results.}} Several groups of experiments are conducted, as mentioned above, and the results are shown in Fig.~\ref{fig:VOT-TIR2015AR-ranking}. It is easy to see that the tracker HSNet achieves better performance than HSNet-lastlayer in terms of accuracy and robustness. Specifically, HSNet improves the accuracy by about 2\% compared to HSNet-lastlayer, which shows that the proposed hierarchical convolutional layer fusion CNN can obtain more robust features of the TIR object than the same CNN using the last layer.
Additionally, it shows that HSNet-ST and HSNet-CA also enhance the performance of the tracking to a large extent. Especially, the spatial transformer improves the accuracy by about 4\% compared to HSNet demonstrating that the spatial transformer is also effective for the tracking task. When we combine the spatial transformer with channel attention, the performance of HSSNet is enhanced by 2\% again. This shows that our spatial-aware network is working.
Finally, the scale estimation strategy is helpful for the accuracy while the robustness exhibits no change, as shown by the A-R plot for the experiment baseline (weighted\_mean) in Fig.~\ref{fig:VOT-TIR2015AR-ranking}.

\subsection{Comparisons with the State-of-the-Art on VOT-TIR 2016}
\label{experiment2}
In this section, we demonstrate that our tracker achieves the favorable results against most state-of-the-art methods on the benchmark VOT-TIR 2016~\cite{VOT-TIR2016}. Additionally, we show some representative tracking results on several challenging sequences.
\begin{figure}[htbp]
\centering
\includegraphics[width=1\textwidth]{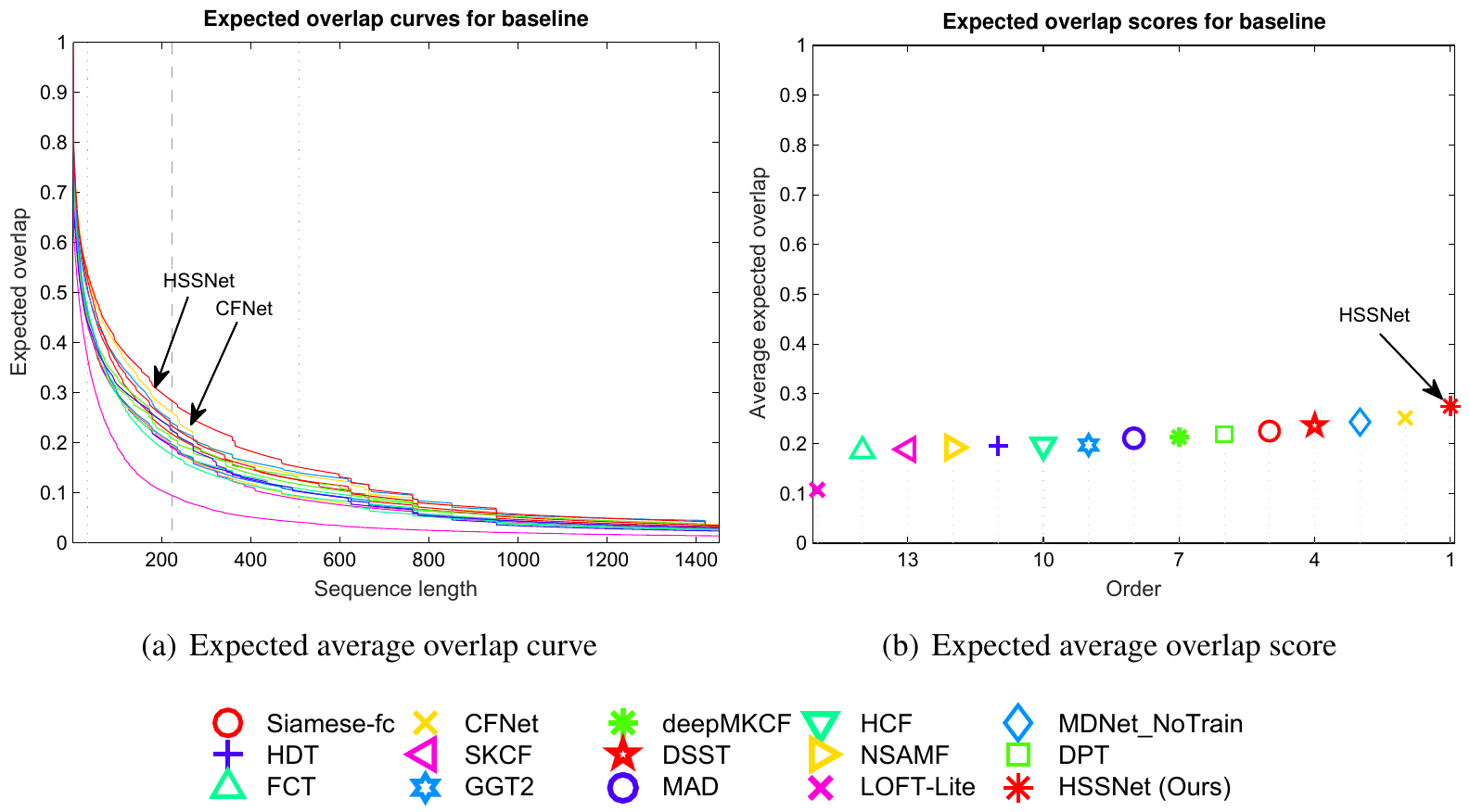}
\caption{EAO curve and EAO score on VOT-TIR 2016. The right-most tracker achieves the best performance according to the EAO values.}
\label{fig:EAO}
\end{figure}

\vspace{3mm}
\noindent{\textbf{Datasets.}} The benchmark VOT-TIR 2016 is an enhanced version of VOT-TIR 2015. Several more challenging sequences have been added.
\vspace{3mm}

\noindent{\textbf{Compared trackers.}} We choose 14 trackers to compare with our tracker on VOT-TIR 2016. These trackers can be divided into four categories: CNN-based trackers, CF-based trackers, part-based trackers, and fusion-based trackers. For the CNN-based trackers, we choose six state-of-the-art methods: Siamese-fc~\cite{Siamese-fc}, MDNet~\cite{MDNet}, CFNet~\cite{CF-NET}, deepMKCF~\cite{DeepMKCF}, HCF~\cite{HCF}, and HDT~\cite{HDT}. These trackers achieve promising results on the object tracking benchmark~\cite{OTB}. For the CF-based trackers, three trackers are selected: DSST~\cite{DSST}, NSAMF~\cite{NSAMF}, and SKCF~\cite{sKCF}, which achieve favorable results on VOT 2014~\cite{VOT2014}. In particular, DSST shows the best performance. Three part-based trackers are chosen for comparison with ours: DPT~\cite{DPT}, FCT~\cite{VOT-TIR2016}, and GGT2~\cite{GGT2}. For the fusion-based trackers,  MAD~\cite{MAD} and LOFT-Lite~\cite{LOFT-Lite} are selected.
\vspace{3mm}

\noindent{\textbf{Evaluation and results.}} Firstly, to evaluate the overall performance of a tracker on the concrete tracking application, the EAO curve and the EAO score have been adopted to visualize it and the results are shown in Fig.~\ref{fig:EAO}. This confirms that our tracker HSNet achieves the best performance.  Specifically, HSSNet has the highest EAO score of $0.2619$, which is about 1\% and 3\% higher than the scores of CFNet~\cite{CF-NET} and DSST~\cite{DSST}, respectively.

\begin{figure*}[ht]
\centering
\includegraphics[width=1\textwidth]{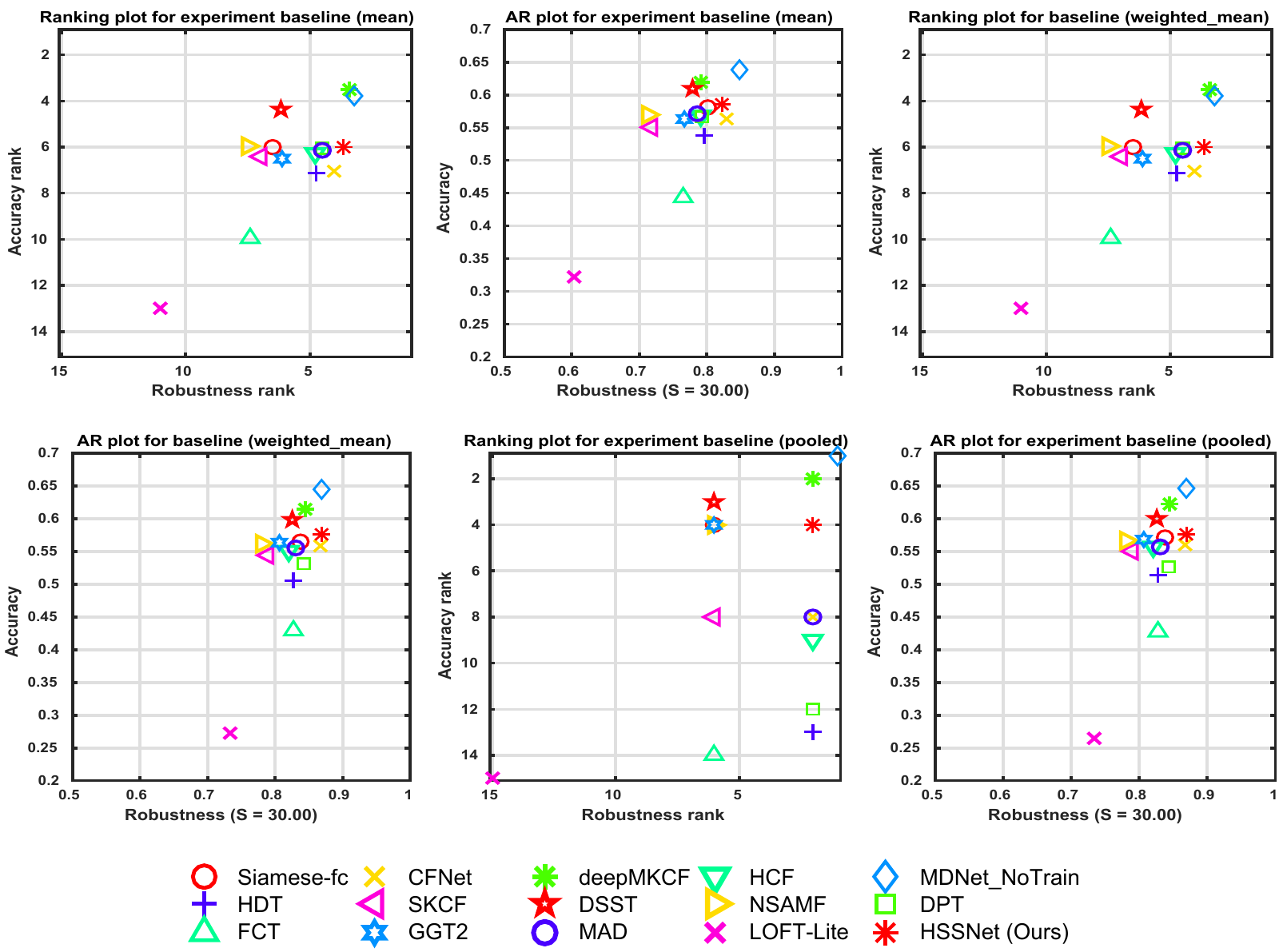}
\caption{A-R ranking plots and the A-R raw plots on VOT-TIR 2016. The better the performance a tracker achieves, the closer to the top right of the plot it appears.}
\label{fig:VOT-TIR2016AR-ranking}
\end{figure*}

Secondly, we show the accuracy and robustness of all these trackers on VOT-TIR 2016, and the results are shown in Fig.~\ref{fig:VOT-TIR2016AR-ranking}. It is obvious that our tracker HSSNet achieves the second best robustness and the fourth best accuracy in all plots. Note that almost all the CNN-based trackers achieve better performance than the conventional hand-crafted features based trackers, which shows that the deep feature is more suitable for TIR tracking, although it is learned from the visible images. Additionally, the CF-based trackers, namely DSST~\cite{DSST}, HCF~\cite{HCF}, and so on, usually perform well in visual tracking while our tracker HSSNet outperforms these trackers in the TIR tracking.

\begin{figure*}[htbp]
\centering
\includegraphics[width=1\textwidth]{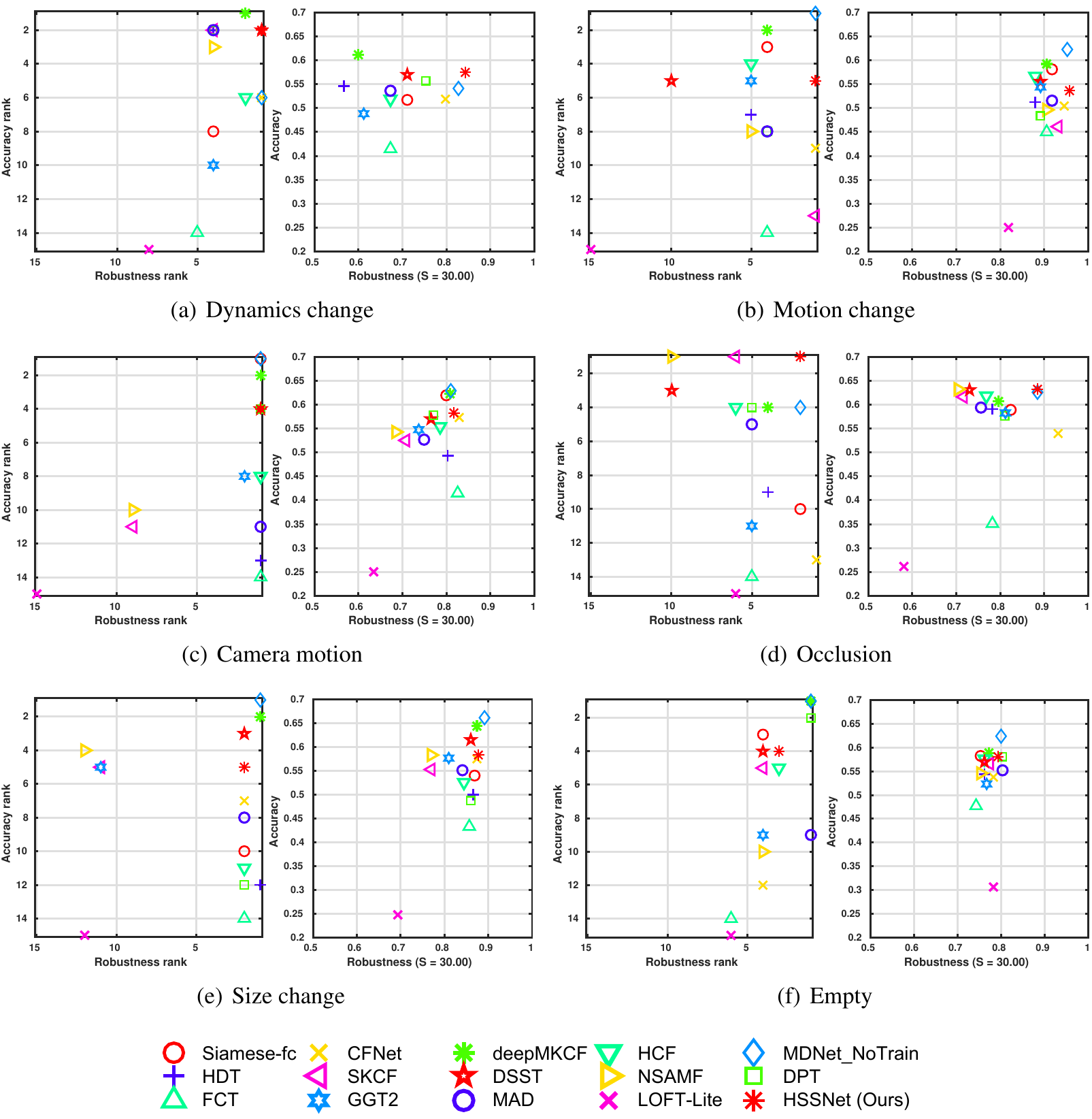}
\caption{A-R ranking plots and the A-R raw plots for the baseline experiments on each attribute subset of VOT-TIR 2016. The better the performance a tracker achieves, the closer to the top-right of the plot it appears.}
\label{fig:VOT-TIR2016attribute}
\end{figure*}

Furthermore, to evaluate the performance of our tracker in handling various challenges , these trackers are compared with our HSSNet on the corresponding attribute subset of VOT-TIR 2016. The results are shown in Fig.~\ref{fig:VOT-TIR2016attribute}. With regard to the dynamics change, motion change, and camera motion challenges, it is easy to see that our method achieves the best robustness, as shown in Figs.~\ref{fig:VOT-TIR2016attribute}a, \ref{fig:VOT-TIR2016attribute}b, and \ref{fig:VOT-TIR2016attribute}c. This demonstrates that our tracker (HSSNet) can deal with these challenges effectively. We suggest that the promising results are obtained from the spatial-aware network, which is robust to the 2D affine transformation.
To the occlusion challenges, it is obvious that our tracker HSSNet achieves the best accuracy and its robustness also exhibits the favorable results, as shown in Fig.~\ref{fig:VOT-TIR2016attribute}d. We suggest that the hierarchical convolutional layer fusion method plays a major role, as it can obtain a richer structure and semantic features for tracking.
{As shown in} Figs.~\ref{fig:VOT-TIR2016attribute}e and \ref{fig:VOT-TIR2016attribute}f, our tracker also obtains a satisfactory result.

\begin{figure*}[htbp]
\centering
\includegraphics[width=1\textwidth]{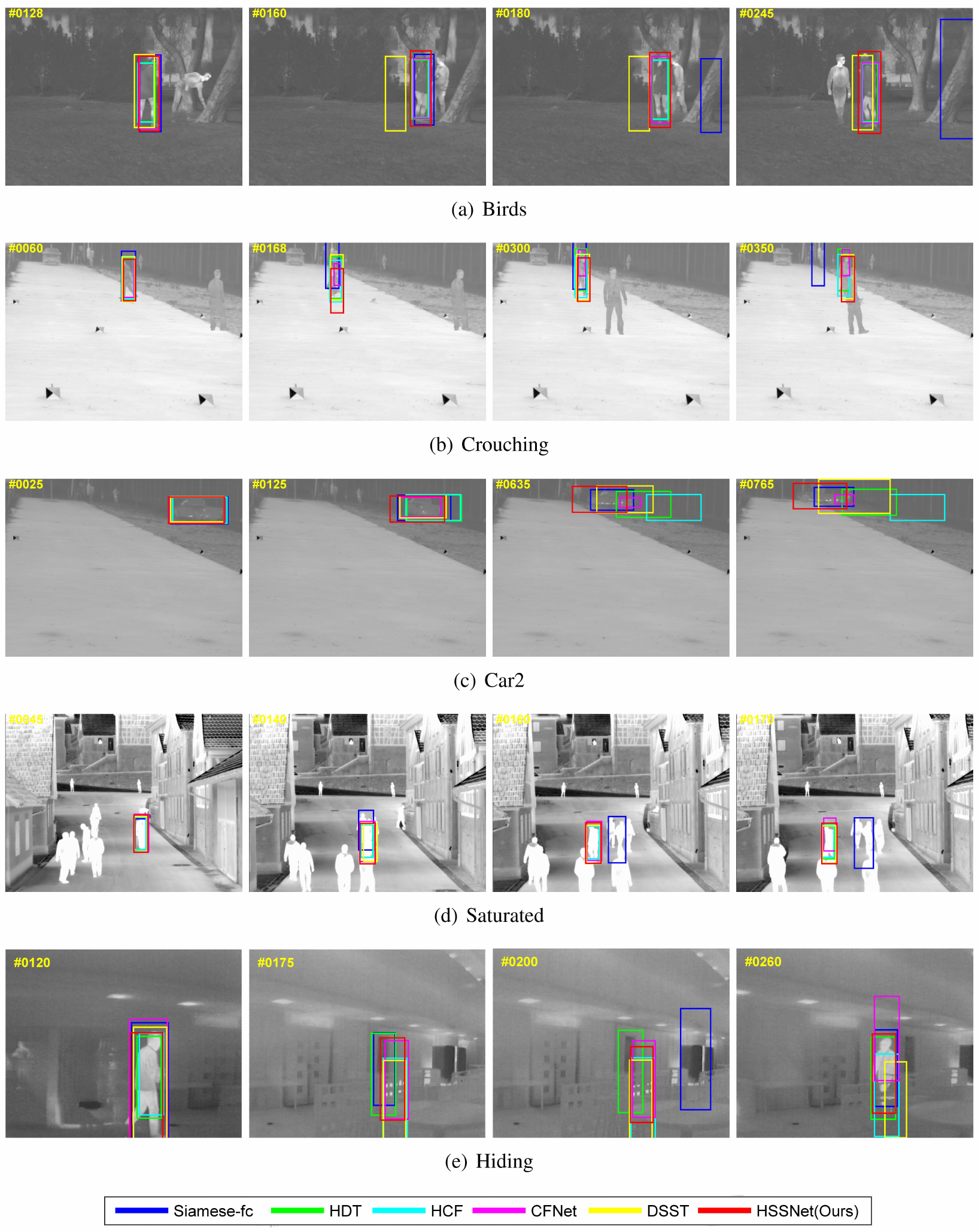}
\caption{Comparison of the visual tracking results of several state-of-the-art trackers on some representative challenging TIR sequences.}
\label{fig:bbresults}
\end{figure*}

Finally, to compare the results of the trackers more intuitively, we give several visual tracking results from the methods evaluated in Fig.~\ref{fig:bbresults}. Overall, it is obvious that our tracker HSSNet locates the targets more precisely. We note that in Fig.~\ref{fig:bbresults}a ("Birds"), DSST and Siamese-fc fail when the target changes its appearance drastically while our tracker tracks the target precisely. In addition, in the "Hiding" sequence, as shown in Fig.~\ref{fig:bbresults}e, we can see that HSSNet outperforms the other trackers when the target is occluded. These results illustrate that HSSNet is robust to variation of the appearance and occlusion challenges. Although the proposed tracker achieves promising results, it often suffers from drift when the background contains heavy noises or very similar objects. The main reason is that the visible image training dataset cannot cover all the real situations of the TIR objects. When two TIR objects have very similar appearances, the proposed network is hard to distinguish them. Therefore, in the future, we plan to make a large-scale TIR training dataset and then fine-tune the proposed network on this dataset to achieve a more promising result.

\section{Conclusion}
\label{Conclusion}
In this paper, {we propose a TIR tracking method (HSSNet) using a hierarchical spatial-aware Siamese CNN. By casting the tracking problem as a similarity verification task, the proposed method achieves the promising results.}
To adapt the TIR tracking, we design a hierarchical convolutional layers fusion method to obtain richer spatial and semantic information. Furthermore, to enhance the discriminative ability of the feature representation, we also propose a spatial-aware network which is good at handling dynamics changes of the object target. Extensive experiments show that our proposed method achieves favorable performance against the state-of-the-art methods.

% use section* for acknowledgment
\section*{Acknowledgment}
This study was supported by the National Natural Science Foundation of China (Grant No. 61672183, U1509216,
61472099, 61502119), by the Natural Science Foundation of Guangdong Province (Grant No. 2015A030313544), by the Shenzhen Research Council (Grant No. JCYJ2017\\0413104556946, JCYJ20170815113552036, JCYJ20160226201453085), and by Shenzhen Medical Biometrics Perception and Analysis Engineering Laboratory.

%% References
%%
%% Following citation commands can be used in the body text:
%% Usage of ~\cite is as follows:
%%   ~\cite{key}         ==>>  [#]
%%   ~\cite[chap. 2]{key} ==>> [#, chap. 2]
%%

%% References with bibTeX database:

\bibliographystyle{elsarticle-num}
% \bibliographystyle{elsarticle-harv}
% \bibliographystyle{elsarticle-num-names}
% \bibliographystyle{model1a-num-names}
% \bibliographystyle{model1b-num-names}
% \bibliographystyle{model1c-num-names}
% \bibliographystyle{model1-num-names}
% \bibliographystyle{model2-names}
% \bibliographystyle{model3a-num-names}
% \bibliographystyle{model3-num-names}
% \bibliographystyle{model4-names}
% \bibliographystyle{model5-names}
% \bibliographystyle{model6-num-names}

%\bibliography{tracking}

\end{document}